\documentclass{article}

\usepackage[square,numbers]{natbib}

\usepackage[preprint]{template_2024}

\usepackage{float}
\usepackage{subfigure}
\usepackage[utf8]{inputenc} %
\usepackage[T1]{fontenc}    %
\usepackage{hyperref}       %
\usepackage{url}            %
\usepackage{booktabs}       %
\usepackage{amsfonts}       %
\usepackage{nicefrac}       %
\usepackage{microtype}      %
\usepackage{xcolor}         %
\usepackage{graphicx} 
\usepackage{listings} %
\usepackage{url}
\usepackage{algorithm}
\usepackage{algorithmic}
\usepackage{amsmath}
\usepackage{amssymb}
\usepackage{amsfonts}
\usepackage{xspace}
\usepackage{subcaption}
\usepackage[subtle]{savetrees}

\usepackage{titlesec}

\titlespacing*{\subsection}{0pt}{1ex plus 1ex minus .2ex}{0.3ex plus .2ex}
\titlespacing*{\section}{0pt}{1.2ex plus 1ex minus .2ex}{0.5ex plus .2ex}

\title{Efficient LLM Scheduling by Learning to Rank}

\author{
Yichao Fu$^{1}$ \quad 
Siqi Zhu$^{2}$ \quad %
Runlong Su$^{1}$ \quad 
Aurick Qiao$^{3}$ \quad 
Ion Stoica$^{4}$ \quad
Hao Zhang$^{1}$\thanks{Hao Zhang is the corresponding author} \\
$^1$UCSD\quad $^2$Tsinghua University\quad $^3$Snowflake\quad $^4$ UC Berkeley \\
}

\begin{document}

\newcommand{\yichao}[1]{{\color{red}{\bf\sf [yichao: #1]}}}
\newcommand{\aurick}[1]{{\color{blue}{\bf\sf [aurick: #1]}}}
\newcommand{\hao}[1]{{\color{blue}{\bf\sf [hao: #1]}}}
\newcommand{\siqi}[1]{{\color{blue}{\bf\sf [siqi: #1]}}}

\newcommand{\maxtpot}[0]{\textsl{max\_waiting\_time}\xspace}

\maketitle

\begin{abstract}
In Large Language Model (LLM) inference, the output length of an LLM request is typically regarded as \emph{not known a priori}. Consequently, most LLM serving systems employ a simple First-come-first-serve (FCFS) scheduling strategy, leading to Head-Of-Line (HOL) blocking and reduced throughput and service quality. 
In this paper, we reexamine this assumption -- we show that, although predicting the exact generation length of each request is infeasible, it is possible to predict the relative ranks of output lengths in a batch of requests, using \emph{learning to rank}. The ranking information offers valuable guidance for scheduling requests. Building on this insight, we develop a novel scheduler for LLM inference and serving that can approximate the shortest-job-first (SJF) schedule better than existing approaches. We integrate this scheduler with the state-of-the-art LLM serving system and show significant performance improvement in several important applications: 2.8x lower latency in chatbot serving and 6.5x higher throughput in synthetic data generation. Our code is available at https://github.com/hao-ai-lab/vllm-ltr.git
\end{abstract}

\section{Introduction}

Large language models (LLMs) are increasingly becoming the backbone of many today's Internet services and applications that serve millions of users~\cite{chatgpt}. Due to the surge in demand, efficient scheduling for LLM serving is crucial to ensure high-quality service amidst numerous concurrent users competing for computing resources. For popular interactive applications such as chatbots, this means minimizing the latency that each user perceives while maximizing the overall system throughput to accommodate as many users as possible.

\begin{figure}[ht]
  \begin{center}
\centerline{\includegraphics[width=0.95\columnwidth]{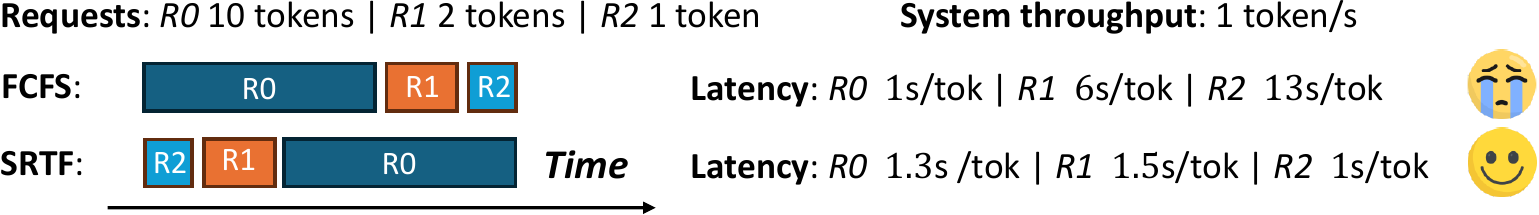}}
\end{center}
\vskip -0.2in
  \caption{A long request can block short requests and introduce severe HOL blocking and high latency. We assume there is no prefill time, and the system takes 1 second to generate 1 token. With a First-come-first-serve (FCFS) schedule, the long request \textit{R0}, which arrives first and takes 10 seconds to generate 10 tokens, will block subsequent shorter requests \textit{R1} and \textit{R2} for 10 seconds. Hence the latencies of \textit{R0},  \textit{R1}, and \textit{R2} are $10 / 10 = 1, (10 + 2) / 2 = 6, (10+2+1)/1=13 \mbox{ s / token}$, respectively, perceived by users, with an average latency of $(1+6+13)/3 = 6.67 \mbox{ s / token}$.
By contrast, prioritizing shortest requests yields an average latency of $(1.3+1.5+1)/3=1.27 \mbox{ s / token}$ -- a $5.3\times$ reduction in average latency.}
  \label{fig:HOL}
\end{figure}

Under high load, LLM services that implement a first-come-first-serve (FCFS) scheduling strategy inevitably face significant Head-Of-Line (HOL) blocking, as many requests must wait for others to execute. Figure~\ref{fig:HOL} illustrates a typical example.  In such scenarios, it is well-established that the shortest-job-first (SJF) and shortest-remaining-time-first (SRTF) scheduling algorithms minimize the average latency experienced across all requests. However, SJF/SRTF are seldom implemented in LLM services because they require requests to be ordered by their remaining generation lengths, which is traditionally assumed to be difficult or impossible to know ahead of time in existing systems~\cite{kwon2023efficient,yu2022orca}.

In this paper, we contend that, although accurately knowing the generation length of requests may be difficult, it is actually not needed. 
Rather, just knowing the \emph{relative ordering} between request lengths is sufficient for SJF/SRTF scheduling.
To this end, we propose to use the Kendall rank correlation coefficient (\emph{Kendall's Tau})~\cite{enwiki:1222948294} to measure the similarity between a predicted schedule and the SJF/SRTF schedule based on groundtruth generation lengths (i.e. oracle). We demonstrate that schedules with higher similarities (measured by Kendal's Tau) to the oracle generally translate to lower latencies in real-world performance (Figure.~\ref{fig:method}). 

Based on this insight, we propose to optimize the request scheduling in LLM serving via learning to rank. We show that a small auxiliary model (e.g., OPT-125M~\cite{zhang2022opt}) can be trained to accurately rank LLM requests by their generation lengths, prior to execution, at virtually no cost. For both offline batch generation and online latency-sensitive tasks, by scheduling requests \emph{on-the-fly} based on the predicted rankings, we can approximate the SRTF/SJF schedule, hence reduce average latency and improve throughput, respectively. 
Compared to existing work which attempts to directly predict the generation lengths of LLM responses~\cite{jin2024s,NEURIPS2023_ce7ff340}, we show that our learning-to-rank approach is both more robust in approximating SRTF/SJF, hence translating to lower latency and higher throughput, but also simpler, which can be easily integrated into production serving systems (i.e., 500 LoC in vLLM). %

Our contributions are summarized as follows:
\begin{itemize}
\item  We show that knowing the relative orderings of generation lengths provides valuable guidance for optimizing the scheduling of LLM serving.
\item  We apply Kendall's Tau as an effective measure of the similarity between an LLM schedule and the ideal SJF/SRTF schedule, and show a higher similiary indicated by Kendall's Tau usually translates to lower latency and high throughput in practice. 
\item We employ \emph{learning-to-rank}~\cite{liu2009learning} to optimize the schedule and show that our method is simple and enables on-the-fly scheduling at a per-iteration basis with negligible overhead.  
\item Our method, when integrated with state-of-the-art serving system, significantly improves the performance on important LLM serving tasks, reducing the p90 latency of chatbot serving by \(2.8\times\) and increasing the throughput of batch synthetic data generation by \(6.5\times\).
\end{itemize}

\section{Related Work}

\textbf{LLM Serving Systems.} Orca~\cite{yu2022orca} introduces iteration-level scheduling and vLLM~\cite{kwon2023efficient} applies PagedAttention, which are two key techniques for LLM serving. However, they both apply the FCFS schedule and are prone to severe HOL blocking. 
Scheduling for LLM serving is a relatively less explored topic. 
Although many LLM serving optimizations~\cite{patel2023splitwise,strati2024d,hu2024inference,zhong2024distserve,agrawal2024taming} have been developed recently, all these works typically assume the output length of an LLM request cannot be known before execution. FastServe~\cite{wu2023fast} applies skip-join MLFQ in LLM serving. It sets up the priority of requests according to their generated length so far. Andes~\cite{liu2024andes} introduces a novel quality of experience (QoE) metric for online text services, which measures human satisfaction during the whole token delivery. It employs an online preemptive scheduling method that determines which requests to execute based on scheduling objectives (e.g., average QoE) for the upcoming timeframe. Our method differs from these by predicting generation length rankings to achieve lower latency.

\textbf{Scheduling in general.} Scheduling is critical in computer systems. First-come-first-serve (FCFS) schedules requests according to their arrival time. Shortest-job-first (SJF) and its preemptive version, shortest-remaining-time-first (SRTF), prioritize jobs with the shortest time to finish, which provably yield the lowest average latency, but may suffer from starvation problems. We discuss how to prevent starvation in \S\ref{sec:starv}.
Multi-level-feedback-queue (MLFQ) maintains multiple priority queues to balance fairness and latency, but introduces substantial complexity in batch and interactive LLM workloads.

\textbf{LLM Generation Length Prediction.} Closest to our work are several recent works that predict the (exact) generation length of LLMs in order to enhance resource utilization (e.g., memory). Perception Only (PO)~\cite{NEURIPS2023_ce7ff340} methods let LLMs output the generation length via prompting. S3~\cite{jin2024s}, TetriServe~\cite{hu2024inference} and DynamoLLM~\cite{stojkovic2024dynamollm} use a predictor model (i.e, DistilBert~\cite{sanh2020distilbert} and OPT~\cite{zhang2022opt}) to predict generation length. These methods formulate the length prediction as a classification problem, whose success hinges on high predictive accuracy. Magnus~\cite{cheng2024enabling} utilizes a language-agnostic BERT sentence embedding, a compression component, and a random forest regressor to predict generation length. Other concurrent works~\cite{qiu2024efficient,qiu2024power} both propose a regression-based method for length prediction, fine-tuning a BERT model on the Lmsys-Chat-1M dataset with an L1 regression loss to predict the exact generation length. They tested models ranging from 300M to 3B and applied various batching policies, including no batching, dynamic batching, and continuous batching, significantly improving latency and throughput under these settings. Additionally, it supports multi-round LLM conversations. In contrast, our proposed method is built on vLLM with paged attention and uses ranking loss to optimize the predictor model. We designed a preemptive scheduling method with starvation prevention to optimize the end-to-end performance of real-world LLM serving systems.

\section{Background}\label{sec:background}

In this section, we introduce several key concepts through the lens of optimizing LLM scheduling.

\begin{figure}
  \centering
  \begin{minipage}{0.25\textwidth}
    \centering
    \includegraphics[width=\linewidth]{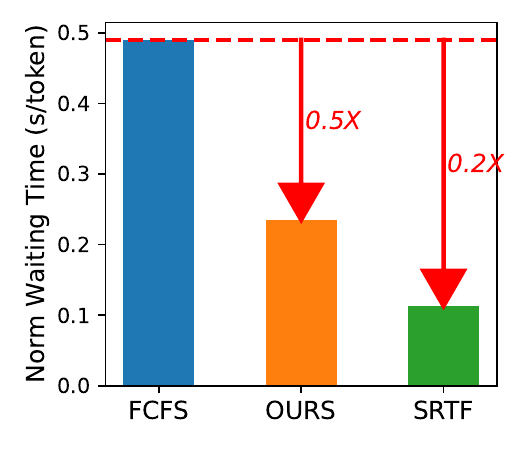} 
    \vspace{-13pt}
    \textbf{(a)}
  \end{minipage}
  \hfill 
  \begin{minipage}{0.7\textwidth}
    \centering
    \includegraphics[width=\linewidth]{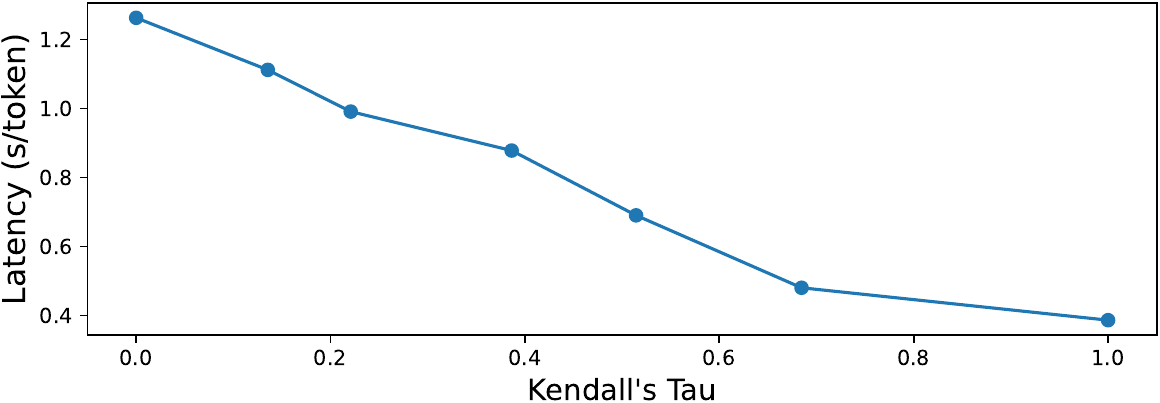}
    \textbf{(b)}
  \end{minipage}
  \vskip -0.05in
  \caption{\textbf{(a)}: HOL blocking of 1K requests on ShareGPT datasets. \textbf{(b)}: Higher Kendall's Tau, lower latency. Evaluated on ShareGPT dataset with Llama-3-8B model. }\label{fig:method}
\end{figure}

\textbf{Kendall Rank Correlation Coefficient.} Kendall's Tau coefficient~\cite{enwiki:1222948294} (we use the Kendall's Tau-b variant) can characterize the correlation between two rankings. Kendall's Tau ranges from \(-1\) to \(1\). \(1\) means two rankings are the same, \(-1\) means two rankings are reversed, and 0 means two rankings are not correlated. The formulation of Kendall's Tau is given as follows:
\begin{equation}
\tau=\frac{N_c-N_d}{\sqrt{\left(N_0-N_1\right)\left(N_0-N_2\right)}},    
\end{equation}
where $N_c$ and $N_d$ are the number of concordant and discordant pairs in two rankings, respectively, $N_0=n(n-1)/2$, $N_1=\sum_i t_i\left(t_i-1\right) / 2$, and $N_2=\sum_j u_j\left(u_j-1\right) / 2$, where $n$ is the total number of items, $t_i$ is the number of tied values in the $i^{th}$ group of ties for the first quantity and $u_j$ is the number of tied values in the $j^{th}$ group of ties for the second quantity~\cite{enwiki:1222948294}. Here, a tied pair is neither concordant nor discordant.

\textbf{Learning to Rank.} Learning to rank~\cite{liu2009learning} applies machine learning methods to ranking supervised data. It is widely used in recommendation systems~\cite{karatzoglou2013learning}, search engine~\cite{liu2009learning} and other research areas~\cite{chen2018learning,chen2024lero}, in three forms: pointwise, pairwise, and listwise. Pointwise turns the ranking problem back to regression~\cite{cossock2006subset}, classification~\cite{li2007mcrank,yin2016ranking} or ordinal regression~\cite{crammer2001pranking}. Pairwise~\cite{freund2003efficient,burges2005learning,zheng2007regression,burges2006learning,wu2010adapting,burges2010ranknet} method learns the relative ranking for each pair. Listwise~\cite{xu2007adarank,taylor2008softrank,cao2007learning,xia2008listwise,pobrotyn2021neuralndcg} learns the ranking of lists of samples in a dataset. 

\textbf{ListMLE.} ListMLE~\cite{xia2008listwise} is a listwise ranking loss of particular interest in our paper. It minimizes the likelihood function defined $\mathcal{\phi}(g(x),y)=-\log P\left(y \mid x ; g\right)$, where
\begin{equation}
P(y \mid x ; g)=\prod_{i=1}^n \frac{\exp \left(g\left(x_{y(i)}\right)\right)}{\sum_{k=i}^n \exp \left(g\left(x_{y(k)}\right)\right)}    
\end{equation}
Here, \( P(y \mid x ; g) \) represents the probability of the permutation \( y \) given the input \( x \) and the scoring function \( g \). \( x_{y(i)} \) denotes the element in \( x \) that corresponds to the \( i \)-th position in the permutation \( y \). The idea is to maximize the likelihood of the correct ranking \( y \) by using the scoring function \( g \) to predict the ranking of the input \( x \). The loss function \( \mathcal{\phi}(g(x),y) \) minimizes the negative log-likelihood of this probability, encouraging the model to predict a ranking close to the true ranking. ListMLE's focus on list ranking aligns with Kendall's Tau, which measures the correlation between two rankings. This ensures that minimizing the loss can help improve Kendall's Tau.

\section{Method}

\subsection{Problem Formulation}

For a given batch of requests, we define the ground truth generation length as $\boldsymbol{l}$, where $\boldsymbol{l_i}$ is the generation length of the $i$-th request in the batch. From this length list, we can obtain a ranking list $\boldsymbol{r}$, where $\boldsymbol{r_i}$ is the rank of $l_i$ in the whole batch $\boldsymbol{l}$.

Our goal is to approximate true SJF/SRTF scheduling using the rankings to alleviate HOL blocking (Fig.~\ref{fig:method} \textbf{a}) and obtain a relatively low latency in LLM serving. Different from the previous methods which target to predict the real generation length $\boldsymbol{l}$, we make predictions on the ranking list $\boldsymbol{r}$. The prediction of the ranking list is defined as $\boldsymbol{p}$ (generated by a predictor $P$). We compute the ranking metric Kendall's Tau~\cite{enwiki:1222948294} to measure the correlation between $\boldsymbol{p}$ and $\boldsymbol{r}$. A Kendall's Tau of 1 means the prediction $\boldsymbol{p}$ perfectly aligns with the ground truth $\boldsymbol{r}$, hence we can use it to achieve perfect SJF/SRTF execution order. A Kendall's Tau of 0 means $\boldsymbol{p}$ is not correlated with $\boldsymbol{r}$. An example is FCFS: the execution order (i.e., by arrival time) is not correlated with the generation length.

A higher Kendall's Tau reflects a more accurate rank prediction against the oracle (i.e., SJF/SRTF), which empirically translates into higher end-to-end performance, as evidenced in Fig.~\ref{fig:method} \textbf{b}. 
Hence, our goal is to optimize the predictor model $P$ to generate predictions with a larger Kendall's Tau, which are more correlated to the ground truth. However, Kendall's tau is inherently non-continuous and difficult to optimize directly. To overcome this, we apply a listwise ranking loss \emph{ListMLE} to optimize the predictor $P$. ListMLE considers the entire list of items simultaneously and treats items at all positions with equal importance, providing a more holistic evaluation of the ranking order compared to other alternatives such as pairwise and pointwise losses.

\subsection{Generation Length Ranking Predictor}\label{sec:pred}

We use a small OPT model as the model backbone to work as the predictor $P$, which can take natural language prompts as input and generate a score for ranking. Previous methods~\cite{NEURIPS2023_ce7ff340,jin2024s,hu2024inference} use classification (with bucketing) to generate the accurate output length predictions, which we find difficult yet unnecessary; instead, the relative ranking suffices. Upon this finding, we apply learning to rank to train this OPT model. We explain how to train the OPT model as the predictor $P$ to rank the prompts by the generation length in this section.

\textbf{Predictor Structure}. The original OPT model can not directly output a score. We append a linear layer to map the hidden states of the last layer to a floating-point number as a score.

\textbf{Training Data}. We aim to train the OPT model to rank the prompts according to their generation length by a target LLM (e.g., Llama-3-70B). So, we need to obtain full generations, and thus the generation length (i.e., the number of the generated tokens),  by feeding the prompts into the target LLM to generate full outputs. In generating model outputs, we sample tokens with a temperature of 1.0, aligning with the evaluation (\S\ref{sec:eval-meth}). The following is an example of the training data. 
\begin{lstlisting}[label=lst:example]
"prompt": "Divide 10 by 4 and remove the remainder.\n"
"output": "\nAnswer: 2 with a remainder of 0."
"output_tokens_length": 12\end{lstlisting}
\vspace{-0.8cm}

After obtaining the generation length, we convert the generation length of sequences to a label presenting the ranking. The simplest way is to rank the generation length directly in the whole training batch and use the ranking as the label for training. We provide insight into the fact that the LLM generation includes some randomness with sampling in real-world serving. So, we bucket the generation lengths by increments of 10 to make the training label more robust to noise. Then, we rank this processed generation length as the training labels.

\textbf{Training}. We train the OPT on 10k samples with a batch size of 32 for 5 epochs. We use the ListMLE loss and the Adam optimizer with a constant learning rate of 2e-5, $\beta_1$ = 0.9, and $\beta_2$ = 0.999. We truncate the prompts to less than 2,048 tokens to satisfy OPT's context length.

Using ranking loss provides several benefits. First, ranking loss focuses on correct ordering rather than precise classification, which is more robust when dealing with a batch of requests where the output length distribution for each bucket is uneven. In contrast, classification loss typically relies on bucket labels for training, which can lead to poor predictive performance for minority buckets in imbalanced datasets. Second, ranking loss ensures a more reasonable distance between related requests, while classification loss makes the predicted probabilities as close to the actual labels as possible. This naturally leads to the pursuit of larger bucket sizes, which is not beneficial for scheduling. Finally, ranking loss can reduce the risk of overfitting. Classification loss forces the model to minimize classification errors on the training requests, which may not generalize well to requests with covariate shifts and cause the model to be highly sensitive to bucket size (see a study in Tab.~\ref{tab:classification}).

\subsection{Request Scheduling with Rankings}\label{sec:starv}

We propose a simple but effective algorithm, detailed in Algorithm~\ref{alg:scheduler}, to schedule requests with the ranking information. The high-level idea is that, for each iteration, we run the predictor model $P$ to score new requests, then \textit{sort} requests according to their generation length rankings and form a running batch according to their orders in the sorted list under the memory or batch size constraints. We incorporated additional mechanisms to prevent starvation of long requests, explained next. Since the ranking-based scheduling algorithm runs at the iteration level, it is compatible with de facto techniques of LLM servings, i.e., continuous batching~\cite{yu2022orca} and PagedAttention~\cite{kwon2023efficient}.

\begin{algorithm}[!h]
\caption{Ranking Scheduler}
\label{alg:scheduler}
\begin{algorithmic}[1]
\STATE {\bfseries Input:} request queue $Q$, predictor model $P$, LLM $M$, hyper-parameter $StarvationThreshold$ prevents request's starvation, hyper-parameter $PriorityQuantum$ limits request's priority time
\WHILE{True}
\STATE Receive batch of new requests $N$ 
\FOR{$r$ {\bfseries in} $N$}
\STATE $r.Score=P(r)$  \COMMENT {\textcolor{violet}{Batch Run Predictor}}
\ENDFOR
\STATE Append $N$ request into $Q$ upon arrival
\STATE $S=Sort(Q)$ according to the pair $(r.Priority, r.Score)$ \COMMENT {\textcolor{violet}{User-defined sort function}}
\STATE $B\leftarrow\emptyset$ \COMMENT {\textcolor{violet}{$B$ is the running batch of the current step}}
\FOR{$r$ {\bfseries in} $S$}
\IF{$B$ is not full}
\STATE $B\leftarrow B + r$
\STATE $r.StarvationCount = 0$ \COMMENT {\textcolor{violet}{Clear $StarvationCount$ is scheduled}}
\IF{$r.Priority$}
\STATE $r.Quantum = r.Quantum - 1$
\ENDIF
\ELSE
\STATE $r.StarvationCount = r.StarvationCount + 1$
\ENDIF
\ENDFOR
\FOR{$r$ {\bfseries in} $Q$}
\IF{$r.StarvationCount\geq StarvationThreshold$} 
\STATE Promote$(r.Priority)$ \COMMENT {\textcolor{violet}{Promote $r$'s priority and assign quantum}}
\STATE $r.StarvationCount=0$
\STATE $r.Quantum=PriorityQuantum$
\ELSIF{$r.Priority$ and $r.Quantum\leq0$}
\STATE Demote$(r.Priority)$ 
\ENDIF
\ENDFOR 

\STATE Execute $B$ with $M$
\STATE Remove finished requests from $Q$ and output
\ENDWHILE

\end{algorithmic}
\end{algorithm}

\textbf{Starvation Prevention.} Executing following SJF/SRTF may lead to starvation for long requests, whose users may wait very long to obtain a response. Different from previous fairness-promoting design~\cite{sheng2023fairness}, which focuses on the fairness between different clients, we propose a \maxtpot fairness metric to evaluate the fairness at per-request level (hence reflecting per-user satisfaction). We define \maxtpot fairness by considering both \emph{Time To First Token} (TTFT) and \emph{Time Per Output Token} (TPOT)~\cite{zhong2024distserve} in LLM serving as follows:

\begin{equation}
\maxtpot=\max(TTFT, \max(TPOT)).
\end{equation}

Intuitively, it characterizes the maximum time interval between receiving two tokens after the user sends a request to the server. A larger \maxtpot indicates that a user needs to wait for a longer time to obtain a response, in other words, a more severe starvation. 

For each scheduling step, we increase the request's starvation count if it is not executed. When a request's starvation count reaches a pre-defined threshold, we will promote this request's priority by allocating ``quantum'' to this request. After running out of allocated ``quantum'', this request will be demoted to the original priority. %
This method prevents starvation at a request level, improves \maxtpot, and ensure user satisfaction. (\S\ref{sec:effec}).

\section{Evaluation}
In this section, we evaluate our proposed method against the baselines and evaluate the effectiveness of each component. We show that our proposed method can achieve state-of-the-art performance in terms of both Kendall's Tau and end-to-end serving performance metrics: latency and throughput. In short, we achieved \(2.8\times\) lower latency in chatbot serving and \(6.5\times\) higher throughput in synthetic data generation. 

\subsection{Evaluation Setup}\label{sec:eval-meth}

\textbf{Testbed.} The end-to-end evaluation testbed is a DGX server consisting of 8 NVIDIA A100 40GB GPUs, 256 vCPUs, and 1TB host memory. GPUs are connected by NVLink.

\textbf{Serving Models.} We use the latest Meta Llama-3 models in two sizes: 8B and 70B~\cite{llama3}. All experiments use FP16/BF16 precision, which is the most common setting in LLM deployment. The 8B model runs on a single GPU, and the 70B model runs on 8 GPUs with tensor parallelism~\cite{shoeybi2020megatronlm}.

\textbf{Workloads.} We evaluate using the ShareGPT~\cite{sharegpt} and LMSYS-Chat-1M~\cite{zheng2023lmsys} datasets, which come from open-ended, real-world conversations with proprietary LLM chatbots such as ChatGPT~\cite{chatgpt} and Claude as well as 25 other open source LLMs. For each dataset and model pair, we sample 10k non-overlapping prompts for serving and 10k for training the ranking predictor. The length distributions of the datasets are provided in Appendix~\ref{sec:dataset}. Model generations are conducted using random sampling with a temperature of 1.0, ensuring consistency during predictor training and serving evaluation, but note our framework is insensitive to the sampling parameters.

\textbf{Evaluation metrics.} For chatbot serving, we measure average and p90 per-token latency, which is the per-request latency divided by the output length. For offline synthetic generation tasks, we use throughput (requests/second) to indicate request generation speed.

\textbf{Scheduler Settings.} We compare our method (i.e., \textbf{ranking predictor}) with four baselines implemented on top of vLLM. Our implementation is based on a recent version of vLLM v0.4.1. 

\begin{itemize}

\item \textbf{FCFS}: We use a FCFS scheduler that supports executing prefill and decode in the same step. For each scheduling step, the scheduler selects requests by earliest arrival time.
\item \textbf{MLFQ}: We implement MLFQ in 1.2k lines of Python codes on vLLM. The MLFQ scheduler leverages chunked prefill from vLLM to run prefill and decode in the same step, as described in FastServe. The implementation's correctness is demonstrated in Appendix~\ref{sec:mlfq}.
\item \textbf{Perception Only (PO)}: We implement Perception Only~\cite{zheng2024response} on vLLM, which lets the LLM itself "say" how many tokens it will generate. We let the LLM generate half of the maximum number of tokens in ~\cite{zheng2024response}(i.e., 30 tokens) to obtain such prediction (i.e., we use 15 tokens) in a FCFS style and extract the generation length information from the generated tokens. After that, we use this generation-length information to optimize serving. %
\item \textbf{Classification}: We train a classifier using OPT model as a backbone. For Llama-3-8B, we use the OPT-125m model, and for Llama-3-70B model, we use the OPT-350m, which can be supported by 8-way tensor parallelism. We follow the setting in S3~\cite{jin2024s} of $number\ of\ buckets=10$ and bucket size of $\frac{max\ context\ length}{ number\ of\ buckets}$ for a high classification accuracy. We map the hidden states of the OPT model to the number of buckets with a linear layer, and use the same training setting as in~\S\ref{sec:pred} but with a cross-entropy loss. 
\item \textbf{Ranking (\textbf{Ours})}: We apply the ranking scheduler (\S\ref{sec:starv}) and use the ranking predictor and training configuration as in (\S\ref{sec:pred}). We use the same size of OPT model as in \textbf{classification}.

\end{itemize}

\subsection{Chatbot Serving Scheduling}\label{eval:chatbot}

\begin{figure}[h]
  \begin{center}
\centerline{\includegraphics[width=\columnwidth]{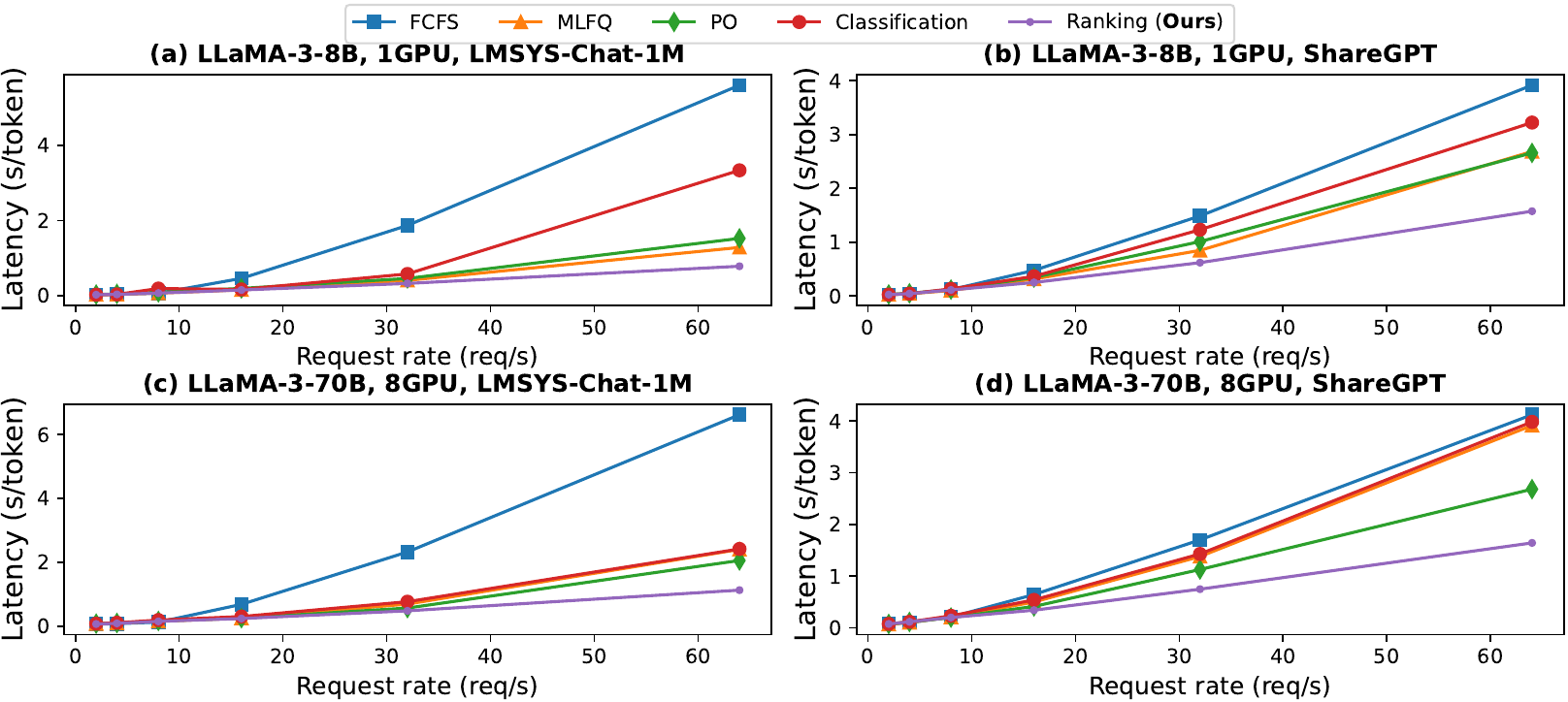}}
\end{center}
\vskip -0.2in
  \caption{Mean latency of different schedulers with Llama-3 models on real workloads.%
  }
  \label{fig:main}
\end{figure}

We compare the latency of the proposed ranking method with four baseline methods on ShareGPT and LMSYS-Chat-1M datasets with increasing arrival rates~\cite{kwon2023efficient,wu2023fast,zhong2024distserve} as in Fig.~\ref{fig:main}. Under the rate of 64 requests/second, our proposed method improve the mean latency by up to \(6.9\times\) compared with FCFS and from \(1.5\times\)--\(1.9\times\) compared with PO. MLFQ and PO face severe HOL blockings as they must run \emph{all} requests for a certain time to obtain information for scheduling. PO must execute all arriving requests with LLM to generate a length prediction. MLFQ must run all arriving requests before they enter the next priority. Classification optimizes towards accuracy instead of ranking, missing optimization opportunities. Classification and our method still need to process all the requests first to obtain a prediction. However, using an OPT model only takes less than 2\% as in~\S\ref{sec:effec}, thus greatly reducing the HOL blocking. 

\noindent \textbf{Handling buristiness}. A burst of submission is a workload in which users submit lots of requests to the LLM server suddenly and is commonly evaluated in previous works~\cite{burst,wang2024towards}. We compare the latency of our method against baselines with a burst of 2k requests as in Tab.~\ref{tab:burst}. Our proposed ranking method can largely improve the latency. We improve the mean latency by up to \(2.0\times\) and improve the P90 latency by up to \(2.8\times\) compared with PO.

\vspace{-0.5cm}

\begin{table}[h]

  \caption{ Latency (s/token) with Burst of 2K requests}
  \label{tab:burst}
  \centering
\scriptsize  
  \begin{tabular}{ll|ccccc|ccccc}
    \toprule
      &   & \multicolumn{5}{c}{Mean Latency (s/token)} & \multicolumn{5}{c}{P90 Latency (s/token)} \\
      \midrule

    Model& Dataset & FCFS & MLFQ & PO & Class. & \textbf{Ours} & FCFS & MLFQ & PO & Class. & \textbf{Ours} \\
    \midrule
   
    Llama-3-8B & ShareGPT & 1.15  & 1.07 & 1.35 & 1.13 & \textbf{0.56} & 1.60  & 1.57 & 1.67 & 1.51 & \textbf{0.67} \\
    Llama-3-8B  &  LMSYS-Chat-1M & 1.73 & 0.80  & 0.75 & 1.77 &  \textbf{0.38}  & 4.86 & 1.56  & 1.47 & 4.98 &  \textbf{0.52} \\
    Llama-3-70B &   ShareGPT  & 1.44 & 1.37 & 1.04 & 1.26 &   \textbf{0.78} & 2.01 & 1.89 & 1.35 & 1.73 &   \textbf{0.96}   \\
    Llama-3-70B &   LMSYS-Chat-1M  & 2.17 & 1.00 & 0.95 & 2.23 & \textbf{0.54}  & 5.54 & 1.91 & 1.72 &  5.72 & \textbf{0.82}  \\

    \bottomrule

  \end{tabular}
\end{table}

\subsection{Synthetic Data Generation Scheduling}\label{eval:synthetic}

Synthetic data generation (SDG) is emerging as an important inference workload due to the data-hungry nature of LLMs. In SDG, we observe sometimes short responses are only preferred for  many practical reasons. First, generating only short conversations is more economical because of the large and diverse number of samples required in SDG~\cite{benallal2024cosmopedia}. Second, long generation leads to evaluation metric bias~\cite{dubois2024alpacafarm,singhal2023long,dubois2024length}. To overcome this, samples with concise generation lengths are preferred for training the model in specific cases. 

Our proposed method can improve the generation throughput in these three cases by preferring short responses. We set up two experiments. 1) we set a quantity limit (i.e., 1k requests) and see how long the schedulers need to generate such generations given 10k prompts. 2) we set a time limit (i.e., 5 minutes) and see how many samples the schedulers can generate given 10k prompts. The results are in  Tab.~\ref{tab:syn-time-1k}. The classification method fails to surpass FCFS because of its extra cost in preprocessing 10k prompts with the OPT model and its low-ranking ability to recognize short requests. Our proposed method, instead, correctly generates short requests by reducing the generation by \(2.4\times\)-\(6.5\times\) compared with FCFS in generating 1k requests and improving the throughput by up to \(3.2\times\) in 5mins.  However, for a setting that does not prefer short generations, the improvement of our algorithm will be minor.

\begin{table}[h]
  \caption{Throughput Improvement with Proposed Ranking Method}
  \label{tab:syn-time-1k}
  \centering
  \scriptsize  
  \begin{tabular}{llcccccc}
    \toprule
      &   & \multicolumn{3}{c}{Time (s) To Generate 1k Samples} & \multicolumn{3}{c}{Generated samples within 5min}  \\
    \cmidrule(r){3-5}  \cmidrule(r){6-8}
    Model  & Dataset  & FCFS & Classification & Ranking \textbf{(Ours)} & FCFS & Classification & Ranking \textbf{(Ours)} \\
    
    \midrule
    Llama-3-8B & ShareGPT & 343.29  & 421.92 & \textbf{143.18} & 841  & 655 & \textbf{1706} \\
    Llama-3-8B  &  LMSYS-Chat-1M & 197.38 &  237.40 & \textbf{30.48}  & 1348 & 1644  & \textbf{4434}   \\
    Llama-3-70B &   ShareGPT  & 440.71 & 512.84 & \textbf{231.59} & 670 & 479 & \textbf{1299}  \\
    Llama-3-70B &   LMSYS-Chat-1M  & 253.68 & 338.83 & \textbf{59.67}  & 1167 & 895 & \textbf{3710} \\

    \bottomrule
  \end{tabular}
\end{table}

\subsection{Comparing Ranking Predictors} 
We show that the accuracy of the targeted classification method is suboptimal in LLM scheduling. We compare the prediction ability of the classification method with different bucket sizes as in Tab.~\ref{tab:classification}. We evaluate the classification metric (i.e., accuracy) for the classification method and the ranking metric (i.e., Kendall's Tau) for all methods on the same randomly sampled test set. A larger bucket size shows better accuracy but does not indicate a higher Kendall's Tau.

We evaluate the end-to-end performance of these methods. Lat. column shows the mean latency to process 2k bursts of requests as in~\S\ref{eval:chatbot}. The Time column shows the time to generate 1k synthetic data as in~\S\ref{eval:synthetic}. A method with a higher Kendall's Tau shows more related with a lower latency, as proposed in~\S\ref{sec:background}. The time to generate 1k synthetic data is less related to Kendall's Tau as a high Tau, but a large bucket size does not mean this predictor can correctly select the shortest requests.

PO achieves higher Kendall's Tau on the LMSYS-Chat-1M dataset. However, it needs to use LLM itself to process all requests and generate a few tokens first for prediction, which introduces a very large HOL overhead compared with light predictor-based methods in spite of its good performance in terms of Kendall's Tau. In all other settings, proposed ranking methods outperform all other methods in terms of ranking metrics and end-to-end performance.   

\textbf{Generalization Ability across Distribution Shifts.} We use the LMSYS-Chat-1M dataset to evaluate the predictor trained on the ShareGPT dataset, and vice versa, to observe its performance under data distribution shifts. The predictor trained on ShareGPT achieves a Kendall's Tau of 0.54 on ShareGPT, and drops to 0.45 when tested on LMSYS-Chat-1M. Conversely, the predictor trained on LMSYS-Chat-1M achieves a Kendall's Tau of 0.62 on LMSYS-Chat-1M, and decreases to 0.40 when tested on ShareGPT.

Although the predictor experiences performance degradation, it still retains some predictive capability, demonstrating a certain level of generalization ability. In real-world scenarios, we can mitigate the impact of distribution shifts by periodically retraining the model with historical data to maintain good ranking prediction performance.

\vspace{-0.4cm}

\begin{table}[h]
\scriptsize  
  \caption{Ranking prediction ability with different classification (Class. in table) settings (i.e., different bucket sizes) for Llama-3-70B. Lat. column shows the mean latency processing a burst of 2k requests for chatbot serving. Time column shows the time to generate 1k requests for synthetic data generation. Optimal Prediction is using the generation length of one random seed to predict the length of another seed. Note that the p-values are below a given significance level (i.e., 1e-3) in all settings.}
  \label{tab:classification}
  \centering
  \begin{tabular}{lcccccccc}
    \toprule
     & \multicolumn{4}{c}{ShareGPT}   & \multicolumn{4}{c}{LMSYS-Chat-1M}                  \\
    \cmidrule(r){2-5}  \cmidrule(r){6-9}
     Method    & Acc. (\%)    & Tau ($\uparrow$) & Lat. (s/tok.) &  Time (s)  & Acc. (\%)    & Tau ($\uparrow$) & Lat. (s/tok.) &  Time (s) \\
    \midrule
    \textit{Optimal Prediction} & / & 0.74 & 0.46  & 102.04 & / & 0.84 & 0.34 & 34.60 \\
    \midrule
    Ranking (\textbf{Ours}) & /  & \textbf{0.54} & \textbf{0.78} & \textbf{231.59}  & / & 0.62 & \textbf{0.54} & \textbf{59.67} \\
    Class. (\#Buckets=10)     & 85.1\% & 0.24 & 1.26 & 512.84 & 96.8\% & 0.17 & 2.23 & 338.83 \\
    Class. (Bucket Size=100)     & 28.1\%       & 0.49 & 0.84 & 265.91 & 43.4\% & 0.58 & 0.77 & 101.61 \\
   
    Class. (Bucket Size=10)     &  4.7\%      & 0.46 & 0.86 & 272.13  & 14.5\% & 0.57 & 0.61 & 78.84 \\
    Class. (Bucket Size=1)     & 1.0\%       & 0.32 & 1.00 & 341.63  & 7.3\% & 0.50 & 0.68 & 92.93 \\
    PO & / & 0.51 & 1.04  & >600 & / & \textbf{0.67} & 0.95 & 322.13 \\

    \bottomrule
  \end{tabular}
\end{table}

\vspace{-0.1cm}

\subsection{Effectiveness Analysis}\label{sec:effec}

\begin{figure}
  \begin{center}
\centerline{\includegraphics[width=0.9\columnwidth]{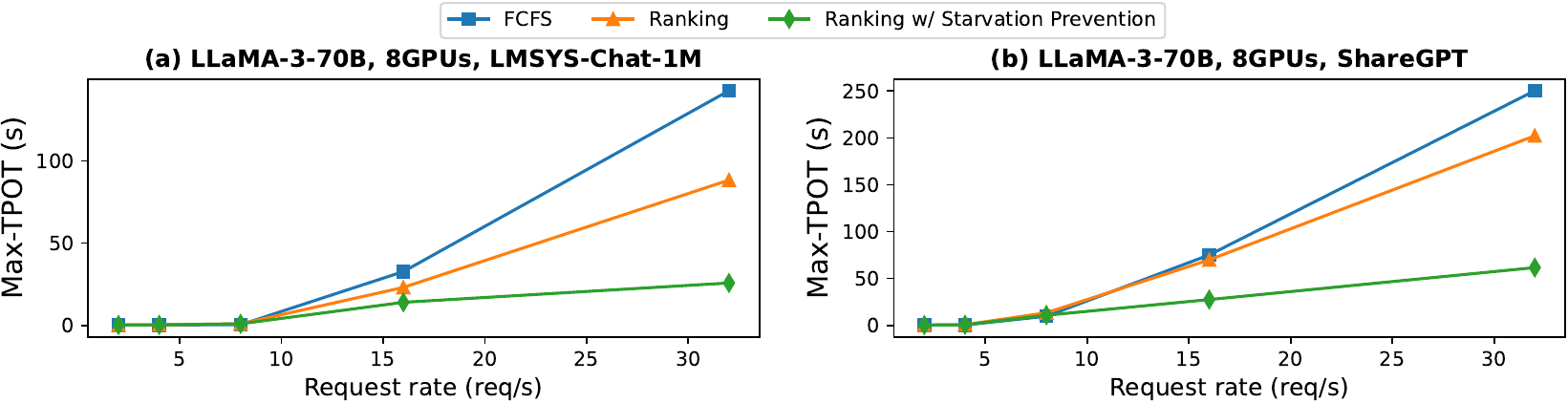}}
\end{center}
\vskip -0.3in
  \caption{Average \maxtpot across all requests with different scheduling method}\label{fig:starv-max-tpot}
\end{figure}

\begin{figure}
  \begin{center}
\centerline{\includegraphics[width=0.9\columnwidth]{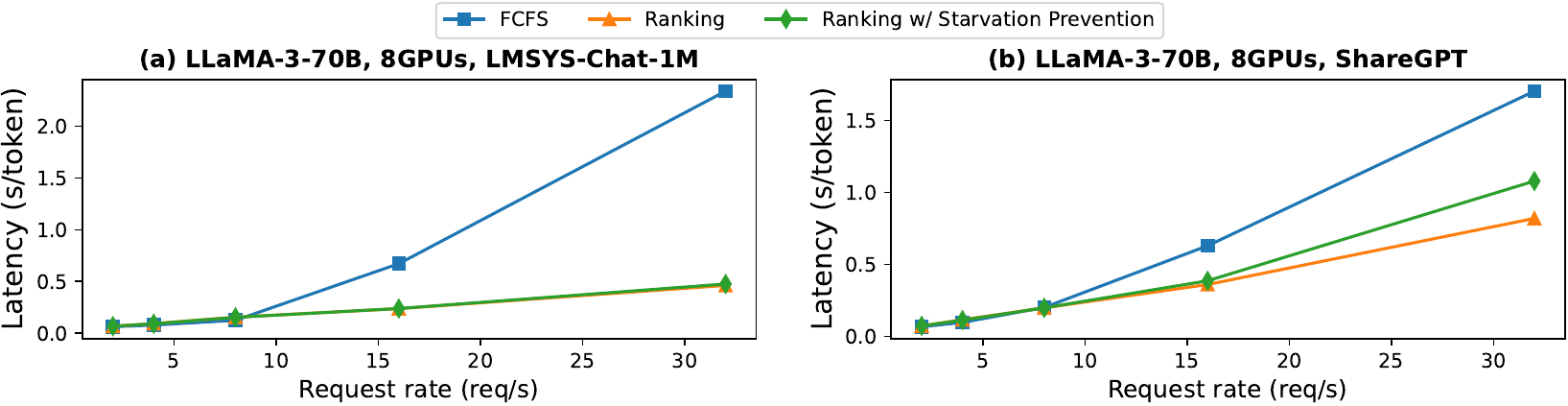}}
\end{center}
\vskip -0.3in
  \caption{Influence of starvation prevention on latency}\label{fig:starv-latency}
\end{figure}

\textbf{Effectiveness of Starvation Prevention.}  We show that our proposed starvation prevention method (\S\ref{sec:starv}) can greatly reduce starvation, decipted by  \maxtpot. The results are shown in Fig.~\ref{fig:starv-max-tpot}. Mean \maxtpot is reduced by up to \(3.4\times\) on LMSYS-Chat-1M and up to \(3.3\times\) on ShareGPT compared with not using starvation prevention. We also show that starvation prevention has little side effects on latency, as in Fig.~\ref{fig:starv-latency}. Starvation prevention reserves latency with less than 10\% overheads in latency in most cases and less than 30\% in all cases, which is an acceptable overhead. 

\vspace{-0.2cm}

\begin{table}[h]
\scriptsize  
  \caption{Overhead of Predictor Model}
  \label{tab:overhead}
  \centering
  \begin{tabular}{llcccc}
    \toprule
    Model  & Dataset  & Overall Time (s) & Prefill Time (s) & Predictor Time (s) & Overhead (\%)  \\
    \midrule
    Llama-3-8B & ShareGPT & 254.23 & 22.34 & 2.81  & 1.11\\
    Llama-3-8B  &  LMSYS-Chat-1M & 127.82 & 7.50 & 1.03 &  0.81     \\
    Llama-3-70B &   ShareGPT  & 419.74 & 46.06 & 7.09 & 1.69 \\
    Llama-3-70B &   LMSYS-Chat-1M  & 211.30 & 15.44 & 2.46  & 1.16  \\

    \bottomrule
  \end{tabular}
\end{table}

\textbf{Overhead of Predictor Model.} We illustrate the overhead of ranking predictor in responsing 1k requests as in Tab.~\ref{tab:overhead}. Prefill Time is measured by only processing the prompts with the original LLM. The overhead of the ranking models (only processing the prompts) is less than 2\% in all settings. The overhead on ShareGPT dataset is slightly higher (i.e., 1.11\% and 1.69\%) because the prompt length of ShareGPT is longer, as in Appendix~\ref{sec:dataset}. The execution time of OPT is 10\%\textasciitilde15\% of the execution time of the original LLM in processing the prompts, largely alleviating the HOL blocking cost by prediction compared with PO in chatbot servings.

\section{Limitations}
\textbf{Limitation of the Ranking Metric.} The ranking metric Kendall's Tau still has limitations in reflecting the performance of end-to-end tasks. For example, assume we have a ranking prediction that correctly reflects the generation length, randomly shuffle the shortest 70\% requests' predictions, and the longest 70\% requests' predictions will give the same Kendall's Tau of 0.5, but the latency will have a difference of \(1.8\times\) for Llama-8B model on ShareGPT dataset. However, for a more uniformly shuffled ranking list in most cases, Kendall's Tau successfully reflects latency as in Fig.~\ref{fig:method}.

\textbf{Limitation of Proposed Ranking Scheduler.} The proposed ranking scheduler currently works with standard LLM serving techniques such as continuous batching and paged attention. How to integrate the scheduler with the latest optimizations, such as 
chunk-prefill~\cite{agrawal2024taming} and prefill-decode disaggregation~\cite{zhong2024distserve}, has still not been fully studied. We will leave them as future work.

\section{Conclusion}

In this paper, we train a predictor to learn the generation length ordering of LLM requests by applying \emph{learning to rank}. We implement a rank scheduler on top of vLLM, and our proposed method shows significant improvement under different tasks: 2.8x lower latency in chatbot serving and 6.5x higher throughput in synthetic data generation. Due to the simplicity and low cost of our method, we believe it can be easily incorporated into production-level LLM serving systems to reduce the serving cost and improve the quality of services.

\newpage
\bibliographystyle{unsrt}
\bibliography{reference}

\begin{thebibliography}{10}

\bibitem{chatgpt}
OpenAI.
\newblock Introducing chatgpt.
\newblock https://openai.com/index/chatgpt/, November 2022.

\bibitem{kwon2023efficient}
Woosuk Kwon, Zhuohan Li, Siyuan Zhuang, Ying Sheng, Lianmin Zheng, Cody~Hao Yu, Joseph Gonzalez, Hao Zhang, and Ion Stoica.
\newblock Efficient memory management for large language model serving with pagedattention.
\newblock In {\em Proceedings of the 29th Symposium on Operating Systems Principles}, pages 611--626, 2023.

\bibitem{yu2022orca}
Gyeong-In Yu, Joo~Seong Jeong, Geon-Woo Kim, Soojeong Kim, and Byung-Gon Chun.
\newblock Orca: A distributed serving system for $\{$Transformer-Based$\}$ generative models.
\newblock In {\em 16th USENIX Symposium on Operating Systems Design and Implementation (OSDI 22)}, pages 521--538, 2022.

\bibitem{enwiki:1222948294}
{Wikipedia contributors}.
\newblock Kendall rank correlation coefficient --- {Wikipedia}{,} the free encyclopedia, 2024.
\newblock [Online; accessed 18-May-2024].

\bibitem{zhang2022opt}
Susan Zhang, Stephen Roller, Naman Goyal, Mikel Artetxe, Moya Chen, Shuohui Chen, Christopher Dewan, Mona Diab, Xian Li, Xi~Victoria Lin, et~al.
\newblock Opt: Open pre-trained transformer language models.
\newblock {\em arXiv preprint arXiv:2205.01068}, 2022.

\bibitem{jin2024s}
Yunho Jin, Chun-Feng Wu, David Brooks, and Gu-Yeon Wei.
\newblock $ s^{3} $: Increasing gpu utilization during generative inference for higher throughput.
\newblock {\em Advances in Neural Information Processing Systems}, 36, 2024.

\bibitem{NEURIPS2023_ce7ff340}
Zangwei Zheng, Xiaozhe Ren, Fuzhao Xue, Yang Luo, Xin Jiang, and Yang You.
\newblock Response length perception and sequence scheduling: An llm-empowered llm inference pipeline.
\newblock In A.~Oh, T.~Naumann, A.~Globerson, K.~Saenko, M.~Hardt, and S.~Levine, editors, {\em Advances in Neural Information Processing Systems}, volume~36, pages 65517--65530. Curran Associates, Inc., 2023.

\bibitem{liu2009learning}
Tie-Yan Liu et~al.
\newblock Learning to rank for information retrieval.
\newblock {\em Foundations and Trends{\textregistered} in Information Retrieval}, 3(3):225--331, 2009.

\bibitem{patel2023splitwise}
Pratyush Patel, Esha Choukse, Chaojie Zhang, {\'I}{\~n}igo Goiri, Aashaka Shah, Saeed Maleki, and Ricardo Bianchini.
\newblock Splitwise: Efficient generative llm inference using phase splitting.
\newblock {\em arXiv preprint arXiv:2311.18677}, 2023.

\bibitem{strati2024d}
Foteini Strati, Sara Mcallister, Amar Phanishayee, Jakub Tarnawski, and Ana Klimovic.
\newblock D$\backslash$'ej$\backslash$avu: Kv-cache streaming for fast, fault-tolerant generative llm serving.
\newblock {\em arXiv preprint arXiv:2403.01876}, 2024.

\bibitem{hu2024inference}
Cunchen Hu, Heyang Huang, Liangliang Xu, Xusheng Chen, Jiang Xu, Shuang Chen, Hao Feng, Chenxi Wang, Sa~Wang, Yungang Bao, et~al.
\newblock Inference without interference: Disaggregate llm inference for mixed downstream workloads.
\newblock {\em arXiv preprint arXiv:2401.11181}, 2024.

\bibitem{zhong2024distserve}
Yinmin Zhong, Shengyu Liu, Junda Chen, Jianbo Hu, Yibo Zhu, Xuanzhe Liu, Xin Jin, and Hao Zhang.
\newblock Distserve: Disaggregating prefill and decoding for goodput-optimized large language model serving.
\newblock {\em arXiv preprint arXiv:2401.09670}, 2024.

\bibitem{agrawal2024taming}
Amey Agrawal, Nitin Kedia, Ashish Panwar, Jayashree Mohan, Nipun Kwatra, Bhargav~S Gulavani, Alexey Tumanov, and Ramachandran Ramjee.
\newblock Taming throughput-latency tradeoff in llm inference with sarathi-serve.
\newblock {\em arXiv preprint arXiv:2403.02310}, 2024.

\bibitem{wu2023fast}
Bingyang Wu, Yinmin Zhong, Zili Zhang, Gang Huang, Xuanzhe Liu, and Xin Jin.
\newblock Fast distributed inference serving for large language models.
\newblock {\em arXiv preprint arXiv:2305.05920}, 2023.

\bibitem{liu2024andes}
Jiachen Liu, Zhiyu Wu, Jae-Won Chung, Fan Lai, Myungjin Lee, and Mosharaf Chowdhury.
\newblock Andes: Defining and enhancing quality-of-experience in llm-based text streaming services.
\newblock {\em arXiv preprint arXiv:2404.16283}, 2024.

\bibitem{stojkovic2024dynamollm}
Jovan Stojkovic, Chaojie Zhang, {\'I}{\~n}igo Goiri, Josep Torrellas, and Esha Choukse.
\newblock Dynamollm: Designing llm inference clusters for performance and energy efficiency.
\newblock {\em arXiv preprint arXiv:2408.00741}, 2024.

\bibitem{sanh2020distilbert}
Victor Sanh, Lysandre Debut, Julien Chaumond, and Thomas Wolf.
\newblock Distilbert, a distilled version of bert: smaller, faster, cheaper and lighter.
\newblock {\em arXiv preprint arXiv:1910.01108}, 2019.

\bibitem{cheng2024enabling}
Ke~Cheng, Wen Hu, Zhi Wang, Peng Du, Jianguo Li, and Sheng Zhang.
\newblock Enabling efficient batch serving for lmaas via generation length prediction.
\newblock {\em arXiv preprint arXiv:2406.04785}, 2024.

\bibitem{qiu2024efficient}
Haoran Qiu, Weichao Mao, Archit Patke, Shengkun Cui, Saurabh Jha, Chen Wang, Hubertus Franke, Zbigniew~T Kalbarczyk, Tamer Ba{\c{s}}ar, and Ravishankar~K Iyer.
\newblock Efficient interactive llm serving with proxy model-based sequence length prediction.
\newblock {\em arXiv preprint arXiv:2404.08509}, 2024.

\bibitem{qiu2024power}
Haoran Qiu, Weichao Mao, Archit Patke, Shengkun Cui, Saurabh Jha, Chen Wang, Hubertus Franke, Zbigniew Kalbarczyk, Tamer Ba{\c{s}}ar, and Ravishankar~K Iyer.
\newblock Power-aware deep learning model serving with $\{$$\mu$-Serve$\}$.
\newblock In {\em 2024 USENIX Annual Technical Conference (USENIX ATC 24)}, pages 75--93, 2024.

\bibitem{karatzoglou2013learning}
Alexandros Karatzoglou, Linas Baltrunas, and Yue Shi.
\newblock Learning to rank for recommender systems.
\newblock In {\em Proceedings of the 7th ACM Conference on Recommender Systems}, pages 493--494, 2013.

\bibitem{chen2018learning}
Tianqi Chen, Lianmin Zheng, Eddie Yan, Ziheng Jiang, Thierry Moreau, Luis Ceze, Carlos Guestrin, and Arvind Krishnamurthy.
\newblock Learning to optimize tensor programs.
\newblock {\em Advances in Neural Information Processing Systems}, 31, 2018.

\bibitem{chen2024lero}
Xingguang Chen, Rong Zhu, Bolin Ding, Sibo Wang, and Jingren Zhou.
\newblock Lero: applying learning-to-rank in query optimizer.
\newblock {\em The VLDB Journal}, pages 1--25, 2024.

\bibitem{cossock2006subset}
David Cossock and Tong Zhang.
\newblock Subset ranking using regression.
\newblock In {\em Learning Theory: 19th Annual Conference on Learning Theory, COLT 2006, Pittsburgh, PA, USA, June 22-25, 2006. Proceedings 19}, pages 605--619. Springer, 2006.

\bibitem{li2007mcrank}
Ping Li, Qiang Wu, and Christopher Burges.
\newblock Mcrank: Learning to rank using multiple classification and gradient boosting.
\newblock {\em Advances in neural information processing systems}, 20, 2007.

\bibitem{yin2016ranking}
Dawei Yin, Yuening Hu, Jiliang Tang, Tim Daly, Mianwei Zhou, Hua Ouyang, Jianhui Chen, Changsung Kang, Hongbo Deng, Chikashi Nobata, et~al.
\newblock Ranking relevance in yahoo search.
\newblock In {\em Proceedings of the 22nd ACM SIGKDD International Conference on Knowledge Discovery and Data Mining}, pages 323--332, 2016.

\bibitem{crammer2001pranking}
Koby Crammer and Yoram Singer.
\newblock Pranking with ranking.
\newblock {\em Advances in neural information processing systems}, 14, 2001.

\bibitem{freund2003efficient}
Yoav Freund, Raj Iyer, Robert~E Schapire, and Yoram Singer.
\newblock An efficient boosting algorithm for combining preferences.
\newblock {\em Journal of machine learning research}, 4(Nov):933--969, 2003.

\bibitem{burges2005learning}
Chris Burges, Tal Shaked, Erin Renshaw, Ari Lazier, Matt Deeds, Nicole Hamilton, and Greg Hullender.
\newblock Learning to rank using gradient descent.
\newblock In {\em Proceedings of the 22nd international conference on Machine learning}, pages 89--96, 2005.

\bibitem{zheng2007regression}
Zhaohui Zheng, Keke Chen, Gordon Sun, and Hongyuan Zha.
\newblock A regression framework for learning ranking functions using relative relevance judgments.
\newblock In {\em Proceedings of the 30th annual international ACM SIGIR conference on Research and development in information retrieval}, pages 287--294, 2007.

\bibitem{burges2006learning}
Christopher Burges, Robert Ragno, and Quoc Le.
\newblock Learning to rank with nonsmooth cost functions.
\newblock {\em Advances in neural information processing systems}, 19, 2006.

\bibitem{wu2010adapting}
Qiang Wu, Christopher~JC Burges, Krysta~M Svore, and Jianfeng Gao.
\newblock Adapting boosting for information retrieval measures.
\newblock {\em Information Retrieval}, 13:254--270, 2010.

\bibitem{burges2010ranknet}
Christopher~JC Burges.
\newblock From ranknet to lambdarank to lambdamart: An overview.
\newblock {\em Learning}, 11(23-581):81, 2010.

\bibitem{xu2007adarank}
Jun Xu and Hang Li.
\newblock Adarank: a boosting algorithm for information retrieval.
\newblock In {\em Proceedings of the 30th annual international ACM SIGIR conference on Research and development in information retrieval}, pages 391--398, 2007.

\bibitem{taylor2008softrank}
Michael Taylor, John Guiver, Stephen Robertson, and Tom Minka.
\newblock Softrank: optimizing non-smooth rank metrics.
\newblock In {\em Proceedings of the 2008 International Conference on Web Search and Data Mining}, pages 77--86, 2008.

\bibitem{cao2007learning}
Zhe Cao, Tao Qin, Tie-Yan Liu, Ming-Feng Tsai, and Hang Li.
\newblock Learning to rank: from pairwise approach to listwise approach.
\newblock In {\em Proceedings of the 24th international conference on Machine learning}, pages 129--136, 2007.

\bibitem{xia2008listwise}
Fen Xia, Tie-Yan Liu, Jue Wang, Wensheng Zhang, and Hang Li.
\newblock Listwise approach to learning to rank: theory and algorithm.
\newblock In {\em Proceedings of the 25th international conference on Machine learning}, pages 1192--1199, 2008.

\bibitem{pobrotyn2021neuralndcg}
Przemys{\l}aw Pobrotyn and Rados{\l}aw Bia{\l}obrzeski.
\newblock Neuralndcg: Direct optimisation of a ranking metric via differentiable relaxation of sorting.
\newblock {\em arXiv preprint arXiv:2102.07831}, 2021.

\bibitem{sheng2023fairness}
Ying Sheng, Shiyi Cao, Dacheng Li, Banghua Zhu, Zhuohan Li, Danyang Zhuo, Joseph~E Gonzalez, and Ion Stoica.
\newblock Fairness in serving large language models.
\newblock {\em arXiv preprint arXiv:2401.00588}, 2023.

\bibitem{llama3}
AI~Meta.
\newblock Introducing meta llama 3: The most capable openly available llm to date.
\newblock {\em Meta AI}, 2024.

\bibitem{shoeybi2020megatronlm}
Mohammad Shoeybi, Mostofa Patwary, Raul Puri, Patrick LeGresley, Jared Casper, and Bryan Catanzaro.
\newblock Megatron-lm: Training multi-billion parameter language models using model parallelism.
\newblock {\em arXiv preprint arXiv:1909.08053}, 2019.

\bibitem{sharegpt}
ShareGPT Team.
\newblock https://sharegpt.com/, 2023.

\bibitem{zheng2023lmsys}
Lianmin Zheng, Wei-Lin Chiang, Ying Sheng, Tianle Li, Siyuan Zhuang, Zhanghao Wu, Yonghao Zhuang, Zhuohan Li, Zi~Lin, Eric Xing, et~al.
\newblock Lmsys-chat-1m: A large-scale real-world llm conversation dataset.
\newblock {\em arXiv preprint arXiv:2309.11998}, 2023.

\bibitem{zheng2024response}
Zangwei Zheng, Xiaozhe Ren, Fuzhao Xue, Yang Luo, Xin Jiang, and Yang You.
\newblock Response length perception and sequence scheduling: An llm-empowered llm inference pipeline.
\newblock {\em Advances in Neural Information Processing Systems}, 36, 2024.

\bibitem{burst}
Cade Daniel, Chen Shen, Eric Liang, and Richard Liaw.
\newblock How continuous batching enables 23x throughput in llm inference while reducing p50 latency.
\newblock https://www.anyscale.com/blog/continuous-batching-llm-inference, June 2023.

\bibitem{wang2024towards}
Yuxin Wang, Yuhan Chen, Zeyu Li, Zhenheng Tang, Rui Guo, Xin Wang, Qiang Wang, Amelie~Chi Zhou, and Xiaowen Chu.
\newblock Towards efficient and reliable llm serving: A real-world workload study.
\newblock {\em arXiv preprint arXiv:2401.17644}, 2024.

\bibitem{benallal2024cosmopedia}
Loubna Ben~Allal, Anton Lozhkov, Guilherme Penedo, Thomas Wolf, and Leandro von Werra.
\newblock Cosmopedia, February 2024.

\bibitem{dubois2024alpacafarm}
Yann Dubois, Chen~Xuechen Li, Rohan Taori, Tianyi Zhang, Ishaan Gulrajani, Jimmy Ba, Carlos Guestrin, Percy~S Liang, and Tatsunori~B Hashimoto.
\newblock Alpacafarm: A simulation framework for methods that learn from human feedback.
\newblock {\em Advances in Neural Information Processing Systems}, 36, 2024.

\bibitem{singhal2023long}
Prasann Singhal, Tanya Goyal, Jiacheng Xu, and Greg Durrett.
\newblock A long way to go: Investigating length correlations in rlhf.
\newblock {\em arXiv preprint arXiv:2310.03716}, 2023.

\bibitem{dubois2024length}
Yann Dubois, Bal{\'a}zs Galambosi, Percy Liang, and Tatsunori~B Hashimoto.
\newblock Length-controlled alpacaeval: A simple way to debias automatic evaluators.
\newblock {\em arXiv preprint arXiv:2404.04475}, 2024.

\bibitem{alpaca}
Rohan Taori, Ishaan Gulrajani, Tianyi Zhang, Yann Dubois, Xuechen Li, Carlos Guestrin, Percy Liang, and Tatsunori~B. Hashimoto.
\newblock Stanford alpaca: An instruction-following llama model.
\newblock \url{https://github.com/tatsu-lab/stanford_alpaca}, 2023.

\end{thebibliography}

\newpage
\appendix
\section{Implementation of MLFQ}\label{sec:mlfq}
\begin{figure}[h]
  \begin{center}
\centerline{\includegraphics[width=0.7\columnwidth]{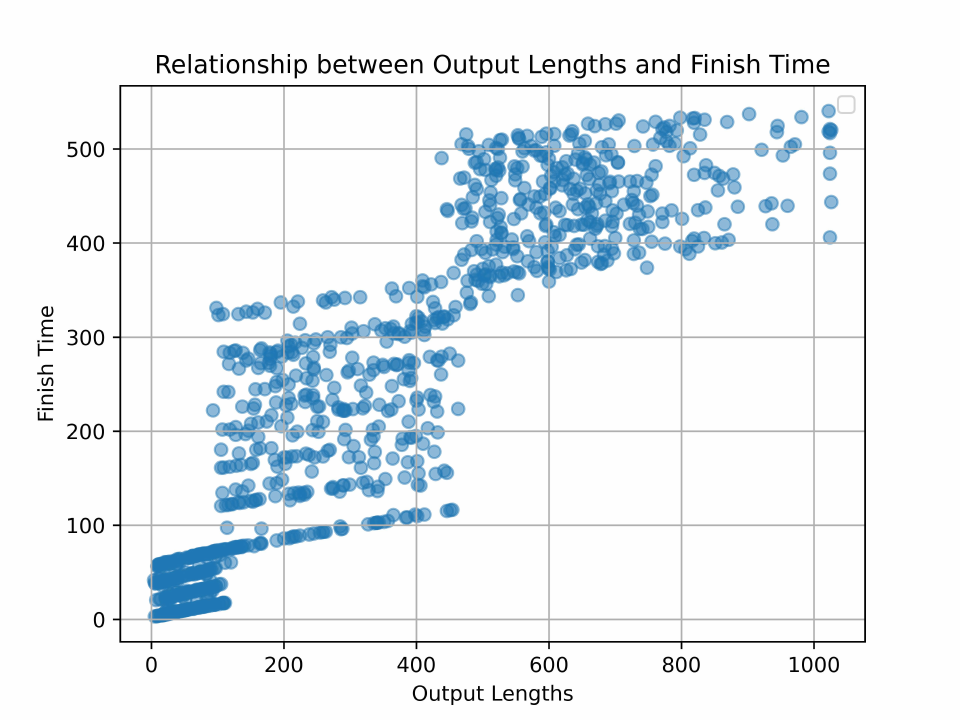}}
\end{center}
\vskip -0.2in
  \caption{Finish Time of Requests with MLFQ Scheduler.}\label{fig:mlfq}
\end{figure}

We are validating the correctness of MLFQ implementation by presenting the relationship of \emph{finish time} and \emph{output length} of requests as shown in Fig.~\ref{fig:mlfq}. This presents a burst of 1k requests with an MLFQ base quantum of 16 seconds, the quantum growth rate of 2, and a max requests limitation of 256 for each step in the vLLM scheduler.

These rectangular blocks, whose edge lengths grow exponentially with the quantum growth rate, represent requests that are completed in queues of varying priorities. When requests from higher priority fail to fill the entire sliding window, those from lower priority begin to be processed, resulting in different blocks being adjacent to one another.

The max request limitation for each step is like a sliding window on all requests. According to the property of MLFQ, requests within the sliding window have two ways out 1) \emph{Finish and pop out} marked by a linear increase in output lengths over time; 2) \emph{Timeout and demote}, occurring when the finish time reaches a multiple of the quantum for the current queue, a batch of requests that arrive at the same time will be demoted simultaneously. With a short quantum for the priority queue, most requests are likely to be demoted rather than completed within the quantum, which explains the clear line trend for the first block shown in the figure. When the finish time reaches multiples of the base quantum (16 seconds in this figure), a new linear growth line appears caused by batch timeout demotions.

\newpage
\section{Dataset Length Distribution }\label{sec:dataset}

\begin{figure}[htbp]
    \centering
    \caption{Dataset Length Distribution}\label{fig:dataset}
    \subfigure
    {
        \begin{minipage}[b]{.48\linewidth}
            \centering
            \includegraphics[scale=0.29]{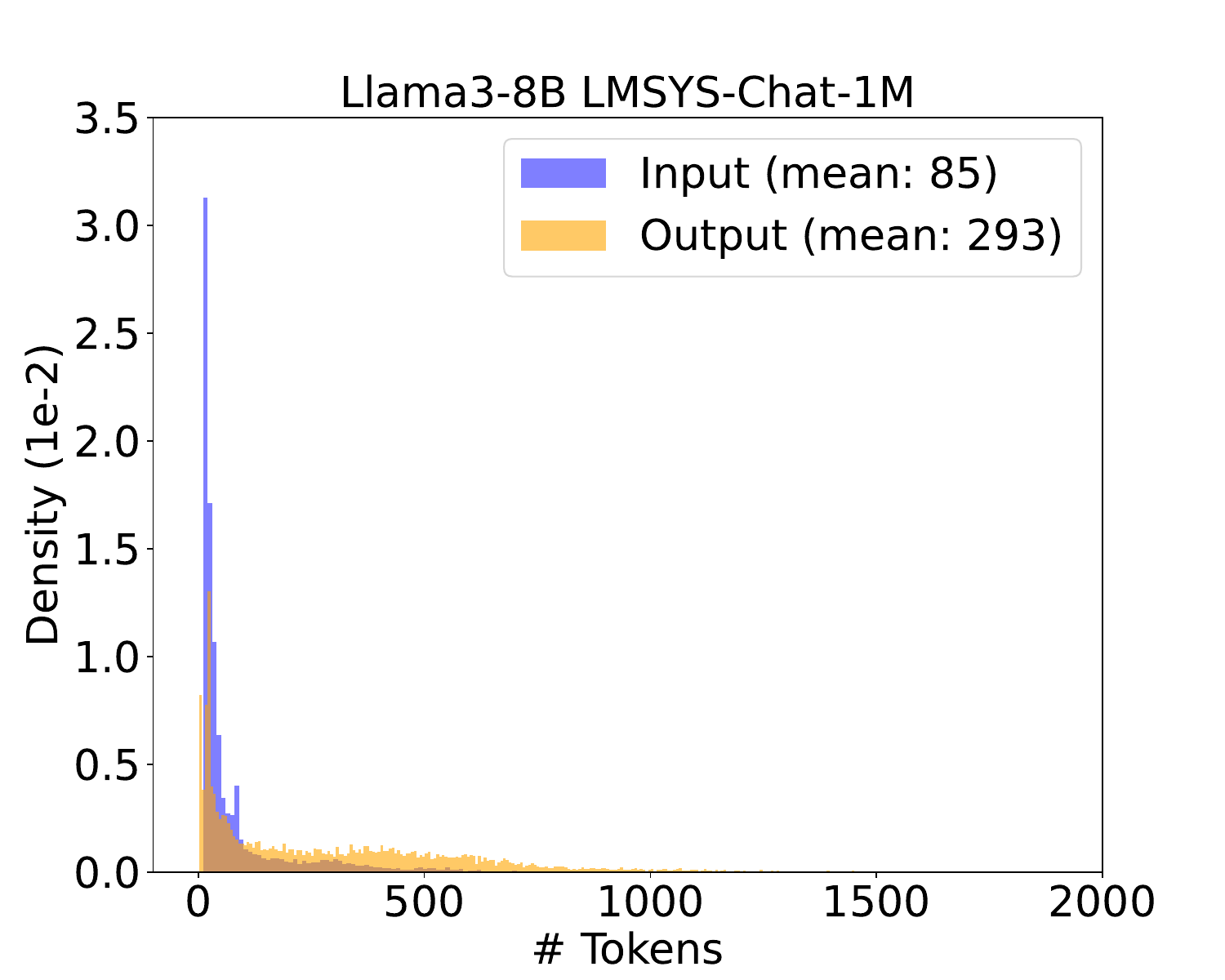}
        \end{minipage}
    }
    \subfigure
    {
        \begin{minipage}[b]{.48\linewidth}
            \centering
            \includegraphics[scale=0.29]{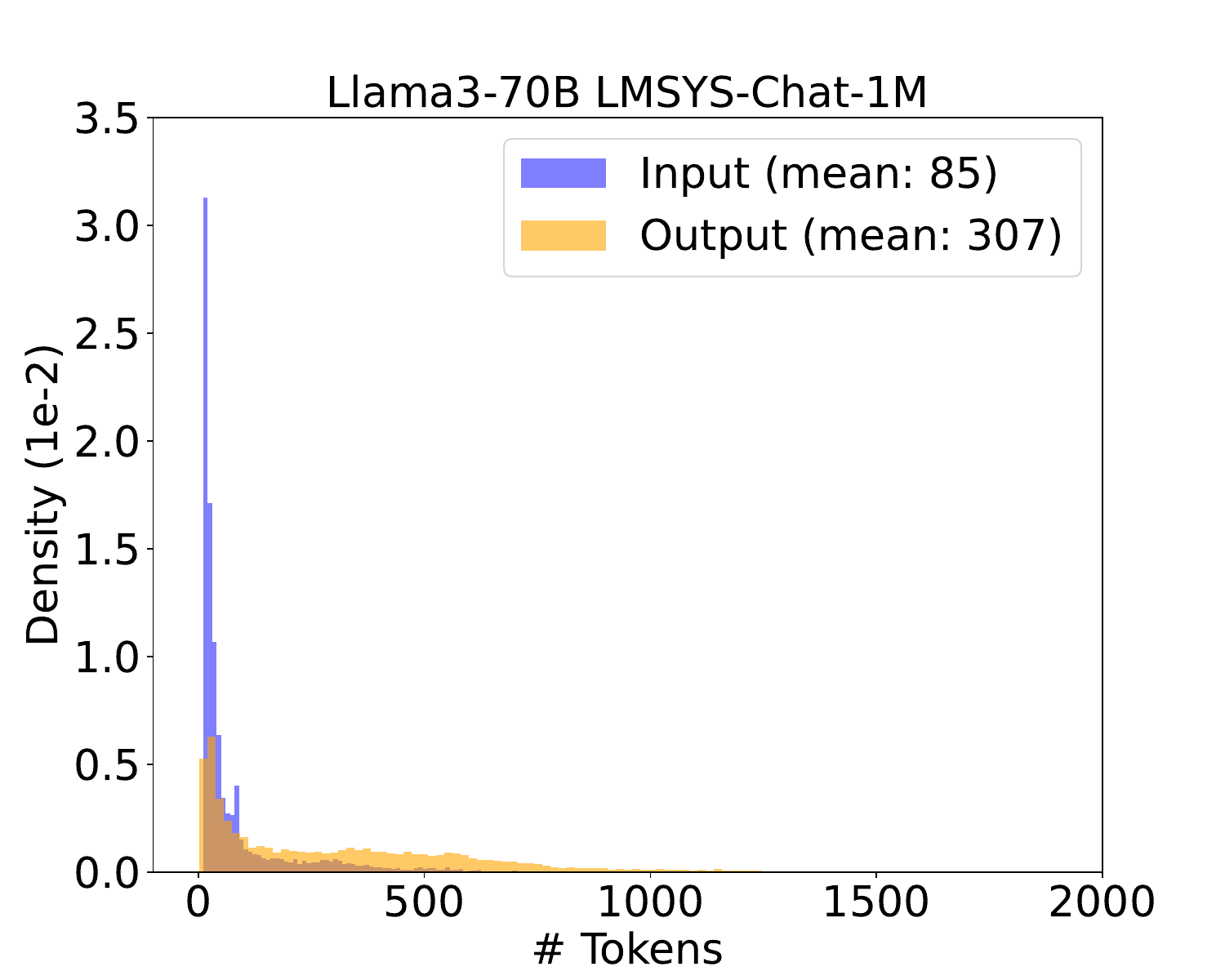}
        \end{minipage}
    }
    \subfigure
    {
        \begin{minipage}[b]{.48\linewidth}
            \centering
            \includegraphics[scale=0.29]{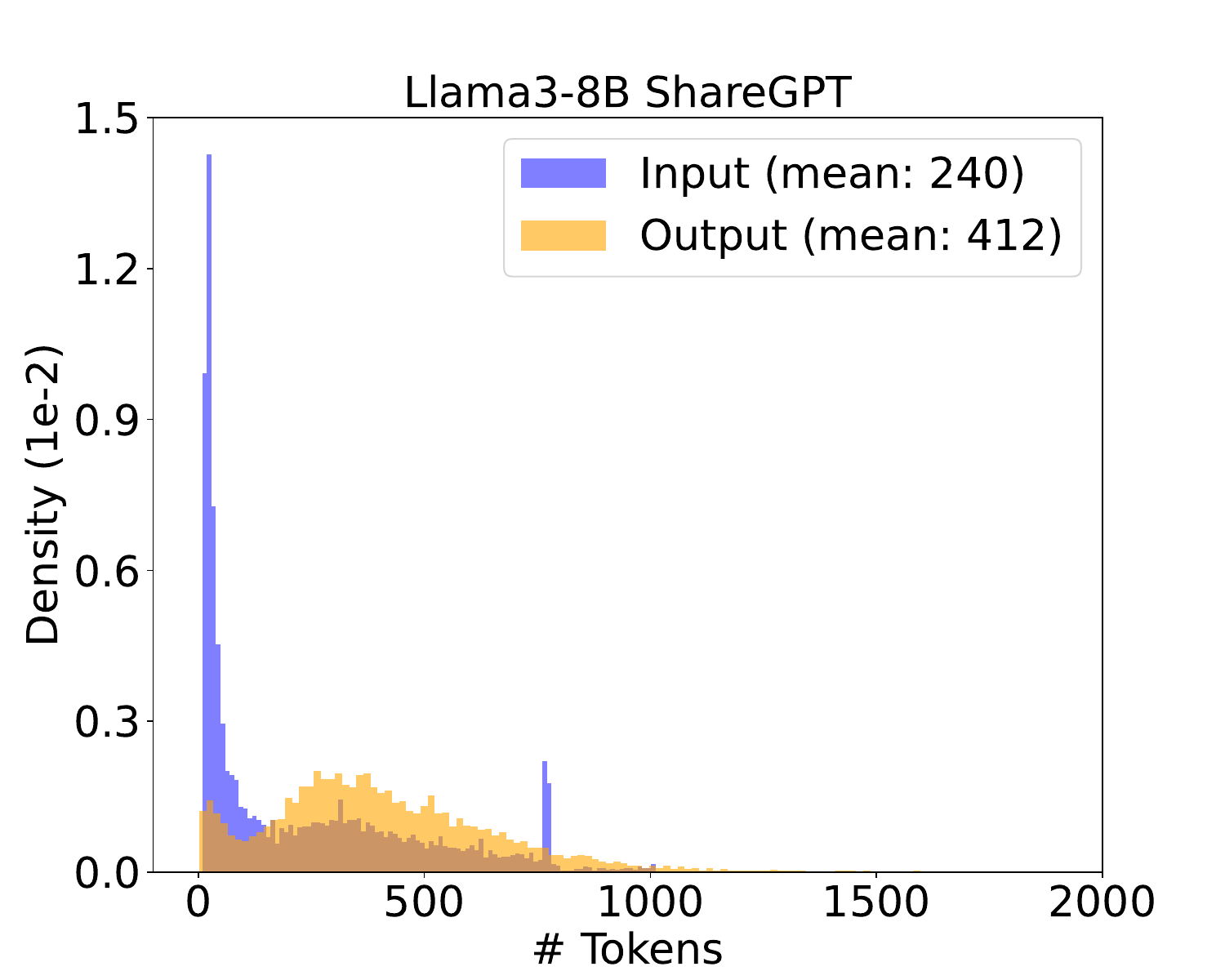}
        \end{minipage}
    } %
    \subfigure
    {
        \begin{minipage}[b]{.48\linewidth}
            \centering
            \includegraphics[scale=0.29]{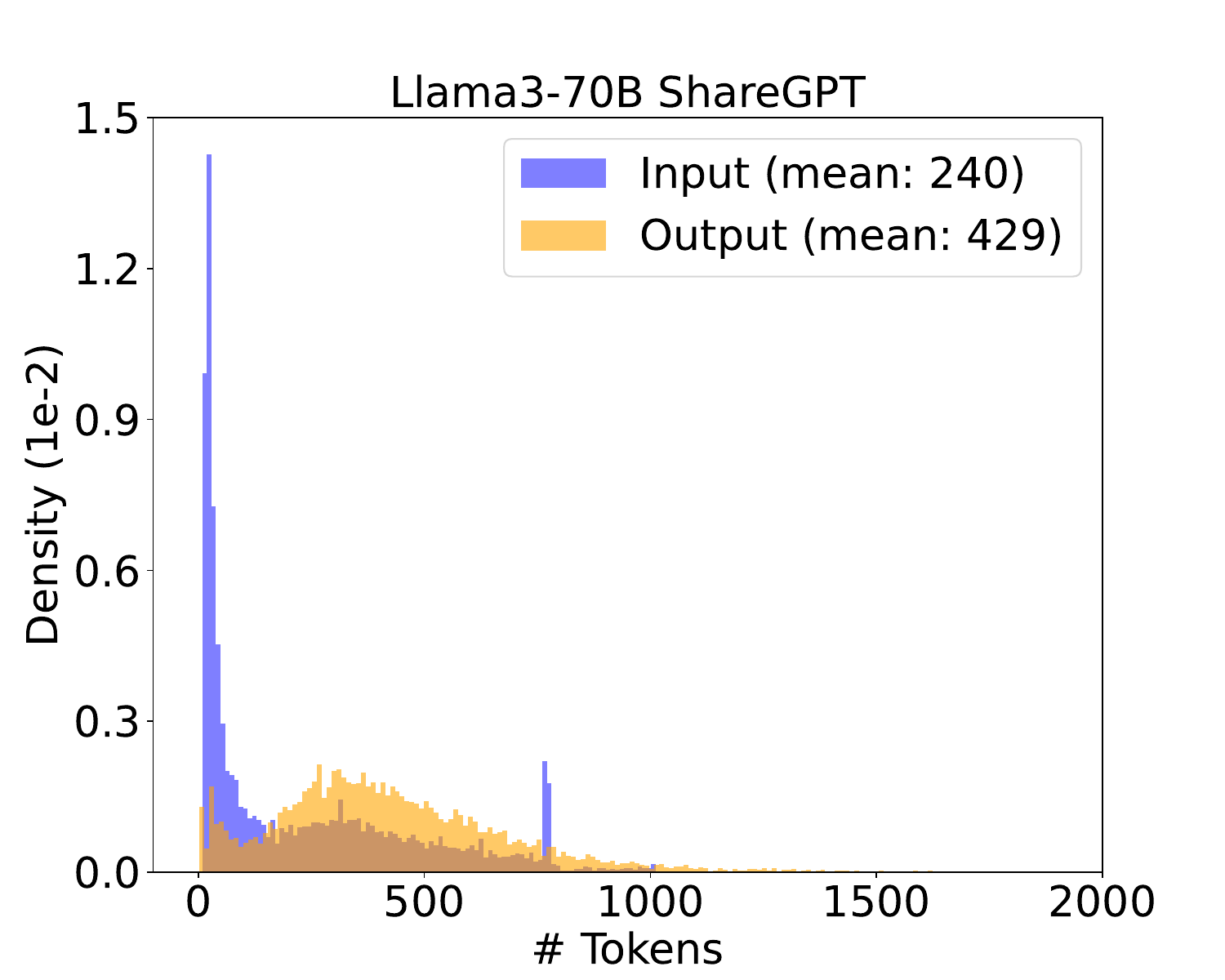}
        \end{minipage}
    }
\end{figure}

We randomly sample 10k samples and present the dataset distribution as in Fig.~\ref{fig:dataset}. We compute the input length by appending the chat template onto the prompts. We have a mean value of 85 input tokens for LMSYS-Chat-1M and a mean value of 240 input tokens for ShareGPT, which is  \(3\times\) longer than LMSYS-Chat-1M. The output length of the ShareGPT dataset is 100 tokens more than the LMSYS-Chat-1M dataset. On average, the 70B version of Llama-3 has a slightly longer output length (i.e., around 15 tokens).

\section{Predictor's Sensitivity to Batch Size}\label{sec:sensitivity}

The ranking scheduler is insensitive to batch size variations. We assess the predictor's sensitivity to batch size on the LMSYS-Chat-1M dataset, as detailed in Tab.~\ref{tab:bs}. We use the predictor to calculate Kendall's Tau for various batch sizes and derive the mean and variance across the entire dataset. This experiment shows that Kendall's Tau remains within a narrow range across different batch sizes. Additionally, our method addresses severe HOL problems when there are numerous requests, in which case, the batch size is often sufficiently large for the predictor to be effective and robust.

\begin{table}[h]
\scriptsize  
  \caption{Predictor's Sensitivity to Batch Size}
  \label{tab:bs}
  \centering
  \begin{tabular}{ccc}
    \toprule
    Batch Size  & Kendall's Tau Mean  & Kendall's Tau Variance \\
    \midrule
    8 & 0.619 & 0.04 \\
    16  & 0.625 & 0.02 \\
    32 &  0.624  & 0.008 \\
    64 &   0.625  & 0.0007 \\
    128 &   0.619  & 0.001 \\
    \bottomrule
  \end{tabular}
\end{table}

\newpage
\section{Relationship Between the ListMLE Loss and the Kendall's Tau}\label{sec:relation}

The ListMLE loss defines a parameterized exponential probability distribution over all scores (as given by the model) and formulates the loss function as the negative log likelihood of the ground truth ranking $y$. Meanwhile, Kendall's Tau measures the ordinal association between the scores (as given by the model) and the ground truth ranking $y$. It is challenging to accurately describe the relationship between the likelihood and ordinal association. However, we provide an analysis demonstrating that minimizing the ListMLE loss can help improve Kendall’s Tau.

To simplify the problem, we assume there are no ties between any two items, meaning each pair should be either concordant or discordant. In this case, Kendall's Tau is defined as $\tau=\frac{N_c-N_d}{n(n-1)/2}$, where $N_c$ and $N_d$ are the number of concordant and discordant pairs in two rankings, and $n$ is the total number of items. As $N_d$ increases, $N_c$ decreases because the sum of $N_c$ and $N_d$ is fixed. Consequently, we have $\Delta \tau=\frac{4\Delta N_c}{n(n-1)}$, where $\tau$ increases when $N_c$ increases.

ListMLE loss is defined as $\mathcal{\phi}(g(x),y)=-\log P\left(y \mid x ; g\right)$, where $P(y \mid x ; g)$ represents the likelihood of the ground truth ranking $y$. As the likelihood of the ground truth ranking $y$ increases, the loss decreases. Although the increase of $P(y \mid x ; g)$ does not guarantee that $N_c$ increases, the increase in the likelihood of the ground truth ranking should generally lead to a greater agreement between the ground truth ranking and the scores given by the model, which implies an increase in the number of concordant pairs (or $N_c$) and a decrease in the number of discordant pairs (or $N_d$) between the scores and the ground truth. Thus, minimizing the loss can help improve Kendall’s Tau.

We further illustrate this relationship by tracking Tau and loss throughout the training process, as shown in Tab.~\ref{tab:rel}. The Pearson correlation coefficient between Tau and loss is -0.9, which means that ListMLE loss and Kendal's Tau coefficient are highly negatively correlated.

\begin{table}[h]
\scriptsize  
  \caption{Relationship Between the ListMLE Loss and the Kendall's Tau}
  \label{tab:rel}
  \centering
  \begin{tabular}{ccc}
    \toprule
    Step  & Kendall's Tau  & Loss \\
    \midrule
    20 & 0.44 & 77.79 \\
    40  & 0.51 & 75.73 \\
    60 &  0.53  & 72.61 \\
    80 &   0.54  & 70.14 \\
    100 &   0.55  & 70.59 \\
    120 &   0.53  & 70.09 \\
    140 &   0.56  & 67.01 \\
    160 &   0.59  & 69.94 \\
    180 &   0.59  & 70.88 \\
    200 &   0.57  & 68.84 \\
    220 &   0.59  & 68.67 \\
    240 &   0.61  & 66.90 \\
    260 &   0.58  & 67.23 \\
    280 &   0.56  & 68.71 \\

    \bottomrule
  \end{tabular}
\end{table}

\newpage
\section{The Performance Gap Between The Proposed Method and Oracle}\label{sec:oracle}

The performance gap between the ranking-based method (ours) and the Oracle varies depending on the evaluation dataset. On certain datasets, our proposed method can perform as well as the Oracle. Due to noise and randomness in the sampling process, we define the Oracle as utilizing sampling results from one seed to guide the scheduling of another sampling, which represents the best performance achievable given one sampling result. For instance, when tested on the Alpaca~\cite{alpaca} dataset with the Llama-8B model, our proposed method closely approximates the Oracle in terms of Kendall's Tau and end-to-end latency for a burst of 2K requests, as depicted in Tab.~\ref{tab:Oracle}. These tests were conducted on a single A100 80G GPU.

On datasets such as LMSYS-Chat-1M and ShareGPT, there remains a small gap between the proposed ranking-based method and the Oracle. The comparison between the ranking-based method (indicated as "Ranking (Ours)") and the Oracle (indicated as "Optimal Prediction") is presented in Tab.~\ref{tab:classification}.

\begin{table}[ht]
\scriptsize  
  \caption{Relationship Between ListMLE Loss and Kendall's Tau}
  \label{tab:Oracle}
  \centering
  \begin{tabular}{ccc}
    \toprule
    & Kendall's Tau  & Latency (s/token) \\
    \midrule
    Ours & 0.73 & 0.28 \\
    Oracle  & 0.72 & 0.24 \\
    FCFS & / &  1.36 \\
    \bottomrule
  \end{tabular}
\end{table}

\section{Influence of The Predictor Size }\label{sec:pred-size}

Our results show that the model size has a minor effect on the prediction ability, as indicated in the following Tab.~\ref{tab:Size}:

The choice to use an OPT-350m model for Llama-70B model is primarily driven by deployment considerations. The OPT-350m model, with 16 attention heads, can be easily deployed using 8-way tensor parallelism, which is also the requirement for the Llama-70B model. In contrast, an OPT-125m model with 12 attention heads cannot be deployed across 8 GPUs, as discussed in \S~\ref{sec:eval-meth}. We deploy the OPT-125m predictor solely on 1 GPU, necessitating the other 7 GPUs to wait when executing the predictor. This configuration results in a waste of resources and may lead to performance degradation.

\begin{table}[h]
\scriptsize  
  \caption{Relationship Between ListMLE Loss and Kendall's Tau}
  \label{tab:Size}
  \centering
  \begin{tabular}{ccc}
    \toprule
    Kendall's Tau  & 125m-OPT & 350m-OPT \\
    \midrule
    ShareGPT & 0.55 & 0.54 \\
    LMSYS-Chat-1M  & 0.64 & 0.62 \\

    \bottomrule
  \end{tabular}
\end{table}

\section{Consideration of Ignoring The Prompt Length}\label{sec:Considers}
In practice, we have found that focusing solely on the generated length is both simple and sufficiently effective.

First, our observations from the Imsys-chat-1M and ShareGPT traces, which represent real-world scenarios, indicate that prompt length is not a critical factor in generation time. Specifically, the prefill time constitutes only 5\% on Imsys-chat-1M and 8\% on ShareGPT, respectively, of the entire generation time, indicating that they have a minor impact on overall latency. Note that there are already long prompts in the workloads we tested. For example, 1\% of all prompts in the ShareGPT dataset exceed 900 tokens.

Second, although this paper does not particularly focus on long contexts (e.g., prompt length > 32k tokens), we argue that handling long prompts is relatively straightforward. Since prompt lengths are always known a priori, it is easy to accurately approximate the latency of the prefill phase through profiling. We can also map the relative ranking of generation length into a length estimation based on the dataset distribution. By simply adding the prefill time estimation to the current framework, we can provide an end-to-end generation time approximation for scheduling.

\newpage
\section{Influence of Correcting Mispredictions Dynamically}\label{sec:Correction}

We have implemented preemptive scheduling, where at each decoding step, we compare the generation rankings of new-incoming requests with those of the currently running requests and preempt those with lower rankings (as detailed in Algorithm ~\ref{alg:scheduler}). However, we do not re-predict the scores for requests that have already been executed during the generation process. Our findings, as presented in Tab.~\ref{tab:corr}, indicate that re-prediction offers minimal improvement. These experiments were conducted using a Llama-3-8B model on a single 80GB A100 GPU.

\begin{table}[h]
\scriptsize  
  \caption{Influence of Correcting Mispredictions}
  \label{tab:corr}
  \centering
  \begin{tabular}{ccc}
    \toprule
    Latency (s/token) & \textbf{Ours} & Re-Prediction \\
    \midrule
    ShareGPT & 0.43 & 0.44 \\
    LMSYS-Chat-1M  & 0.64 & 0.64 \\

    \bottomrule
  \end{tabular}
\end{table}

\end{document}